\documentclass{llncs}

\usepackage[linesnumbered,lined,boxed,vlined]{algorithm2e}
\usepackage{amsfonts}
\usepackage{amsmath}
\usepackage{arydshln}  
\usepackage[table]{xcolor}  
\usepackage{graphics}  
\usepackage{multirow}  
\def\b{\ensuremath\boldsymbol}
\usepackage{graphicx}

\usepackage{amssymb}

\usepackage{floatrow}
\newfloatcommand{capbtabbox}{table}[][\FBwidth]

\usepackage{url}
\usepackage{tikz}
\usepackage{lipsum}
\newcommand\copyrighttextt{%
  \footnotesize Accepted (to appear) in Canadian Conference on Artificial Intelligence (Canadian AI conference) 2020, Springer. This version includes supplementary material for derivation of an equation.}
\newcommand\copyrightnotice{%
\begin{tikzpicture}[remember picture,overlay]
\node[anchor=south,yshift=10pt] at (current page.south) {\fbox{\parbox{\dimexpr\textwidth-\fboxsep-\fboxrule\relax}{\copyrighttextt}}};
\end{tikzpicture}%
}

\begin{document}

\title{Anomaly Detection and Prototype Selection Using Polyhedron Curvature}
%


\author{Benyamin Ghojogh, Fakhri Karray, Mark Crowley
}

\institute{Electrical and Computer Engineering, University of Waterloo, Waterloo, ON, Canada  \\
\email{\{bghojogh, karray, mcrowley\}@uwaterloo.ca}
}

\maketitle              

\begin{abstract}
We propose a novel approach to anomaly detection called Curvature Anomaly Detection (CAD) and Kernel CAD based on the idea of polyhedron curvature. Using the nearest neighbors for a point, we consider every data point as the vertex of a polyhedron where the more anomalous point has more curvature. We also propose inverse CAD (iCAD) and Kernel iCAD for instance ranking and prototype selection by looking at CAD from an opposite perspective. We define the concept of anomaly landscape and anomaly path and we demonstrate an application for it which is image denoising. The proposed methods are straightforward and easy to implement. Our experiments on different benchmarks show that the proposed methods are effective for anomaly detection and prototype selection.
\end{abstract}

\copyrightnotice

\section{Introduction}

Anomaly detection, instance ranking, and prototype selection are important tasks in data mining.
Anomaly detection refers to finding outliers or anomalies which differ significantly from the normal data points \cite{emmott2013systematic}. There exist many applications for anomaly detection such as fraud detection, intrusion detection, medical diagnosis, and damage detection \cite{chandola2009anomaly}.

Ranking data points (instances) according to their importance can be useful for better representation of data, omitting the dummy or noisy points, better discrimination of classes in classification tasks, etc \cite{ghojogh2018principal}. Prototype selection is referred to finding the best data points in terms of representation of data, discrimination of classes, information of points, etc \cite{garcia2012prototype}. It can also be useful for better storage and processing time efficiency. Prototype selection can be done either using ranking the points and then discarding the less important points or by merely retaining a portion of data and discarding the others. 

In this paper, we propose Curvature Anomaly Detection (CAD) and inverse CAD (iCAD) for anomaly detection and prototype selection, respectively. We also propose their kernel versions which are Kernel CAD (K-CAD) and Kernel iCAD (K-iCAD). The idea of proposed algorithms is based on polyhedron curvature where every point is imagined to be the vertex of a hypothetical polyhedron defined by its neighbors. We also define anomaly landscape and anomaly path which can have different applications such as image denoising. 
In the following, we mention the related work for anomaly detection and prototype selection. Then, we explain the background for polyhedron curvature. Afterwards, the proposed CAD, K-CAD, iCAD, and K-iCAD are explained. Finally, the experiments are reported. 


\textbf{Anomaly Detection:} Local Outlier Factor (LOF) \cite{breunig2000lof} is one of the important anomaly detection algorithms. It defines a measure for local density of every data point according to its neighbors. It compares the local density of every point with its neighbors and find the anomalies. 
One-class SVM \cite{scholkopf2000support} is another method which estimates a function which is positive on the regions of data with high density and negative elsewhere. Therefore, the points with negative values of that function are considered as anomalies. 
If the data are assumed to have Gaussian distribution as the most common distribution, Elliptic Envelope (EE) can be fitted to data \cite{rousseeuw1999fast} and the points having low probability in the fitted envelope are considered to be anomaly.
Isolation forest \cite{liu2008isolation} is an isolation-based anomaly detection method \cite{liu2012isolation} which isolates the anomalies using an ensemble approach. The ensemble includes isolation trees where the more depth of tree for isolating a point is a measure of its normality.


\textbf{Prototype Selection:} Prototype selection \cite{garcia2012prototype} is also referred to as instance ranking and numerosity reduction. 
Edited Nearest Neighbor (ENN) \cite{wilson1972asymptotic} is one of the oldest prototype selection method which removes the points having most of its neighbors from another class.  
Decremental Reduction Optimization Procedure 3 (DROP3) \cite{wilson2000reduction} has the opposite perspective and removes a point if ite removal improves the $k$-Nearest Neighbor ($k$-NN) classification accuracy.  
Stratified Ordered Selection (SOS) \cite{kalegele2012demand} starts with boundary points and then recursively finds the median points noticing that boundary and median points are informative. 
Shell Extraction (SE) \cite{liu2017efficient} introduces a reduction sphere and removes the points falling in this hyper-sphere in order to approximate the support vectors. 
Principal Sample Analysis (PSA) \cite{ghojogh2018principal}, which is extended for regression and clustering tasks in \cite{ghojogh2019principal}, considers the scatter of data as well as the regression of prototypes for better representation. 
Instance Ranking by Matrix Decomposition (IRMD) \cite{ghojogh2019instance} decomposes the matrix of data and makes use of the bases of decomposition. The more similar points to the bases are considered to be more important.

\section{Background on Polyhedron Curvature}

\begin{figure}[!t]
\centering
\includegraphics[width=2.7in]{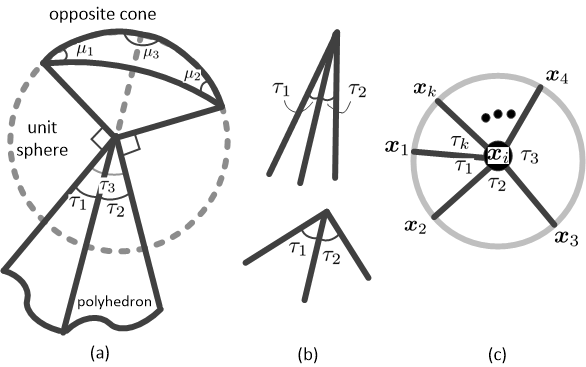}
\caption{(a) Polyhedron vertex, unit sphere, and the opposite cone, (b) large and small curvature, (c) a point and its neighbors normalized on a unit hyper-sphere around it.}
\label{figure_polyhedron}
\end{figure}

A \textit{polytope} is a geometrical object in $\mathbb{R}^d$ whose faces are planar. The special cases of polytope in $\mathbb{R}^2$ and $\mathbb{R}^3$ are called \textit{polygon} and \textit{polyhedron}, respectively. 
Some examples for polyhedron are cube, tetrahedron, octahedron, icosahedron, and dodecahedron with four, eight, and twenty triangular faces, and twelve flat faces, respectively \cite{coxeter1973regular}.
Consider a polygon where $\tau_j$ and $\mu_j$ are the interior and exterior angles at the $j$-th vertex; we have $\tau_j + \mu_j = \pi$.
A similar analysis holds in $\mathbb{R}^3$ for Fig. \ref{figure_polyhedron}-a. In this figure, a vertex of a polyhedron and its opposite cone are shown where the opposite cone is defined to have perpendicular faces to the faces of the polyhedron at the vertex. The intersection of a unit sphere centered at the vertex and the opposite cone is shown in the figure.
This intersection is a geodesic on the unit sphere.
According to Thomas Harriot's theorem proposed in 1603 \cite{markvorsen1996curvature}, if this geodesic on the unit sphere is a triangle, its area is $\mu_1 + \mu_2 + \mu_3 - \pi = 2\pi - (\tau_1 + \tau_2 + \tau_3)$. 
The generalization of this theorem from a geodesic triangular polygon ($3$-gon) to an $k$-gon is $\mu_1 + \dots + \mu_k - k \pi + 2\pi = 2\pi - \sum_{a=1}^k \tau_a$ \cite{markvorsen1996curvature}, where the polyhedron has $k$ faces meeting at the vertex. 

The Descartes's \textit{angular defect} at a vertex $\b{x}$ of a polyhedron is \cite{descartes1890progymnasmata}: $\mathcal{D}(\b{x}) := 2\pi - \sum_{a=1}^k \tau_a$.
The total defect of a polyhedron is defined as the summation of the defects over the vertices.
It can be shown that the total defect of a polyhedron with $v$ vertices, $e$ edges, and $f$ faces is: $\mathcal{D} := \sum_{i=1}^v \mathcal{D}(\b{x}_i) = 2\pi (v - e + f)$.
The term $v - e + f$ is Euler-Poincar{\'e} characteristic of the polyhedron \cite{richeson2019euler,hilton1982descartes}; therefore, the total defect of a polyhedron is equal to its Euler-Poincar{\'e} characteristic. 
According to Fig. \ref{figure_polyhedron}-b, the smaller $\tau$ angles result in sharper corner of the polyhedron. Therefore, we can consider the angular defect as the \textit{curvature} of the vertex. 

\section{Anomaly Detection}

\subsection{Curvature Anomaly Detection}

The main idea of the \textit{Curvature Anomaly Detection (CAD)} method is as follows. 
Every data point is considered to be the vertex of a hypothetical polyhedron (see Fig. \ref{figure_polyhedron}-a). 
For every point, we find its $k$-Nearest Neighbors ($k$-NN).
The $k$ neighbors of the point (vertex) form the $k$ faces of a polyhedron meeting at that vertex. 
Then, the more curvature that point (vertex) has, the more anomalous it is because it is far away (different) from its neighbors. 
Therefore, \textit{anomaly score} $s_A$ is proportional to the curvature. 

Since, according to the equation of angular effect, the curvature is proportional to minus the summation of angles, we can consider the anomaly score to be inversely proportional to the summation of angles. Without loss of generality, we assume the angles to be in range $[0, \pi]$ (otherwise, we take the smaller angle). The less the angles between two edges of the polyhedron, the more their cosine. 
As the anomaly score is inversely proportional to the angles, we can use cosine for the anomaly score: $s_A(\b{x}_i) \propto 1 / \tau_a \propto \cos(\tau_a)$.
We define the anomaly score to be the summation of cosine of the angles of the polyhedron faces meeting at that point: $s_A(\b{x}_i) := \sum_{a=1}^{k} \cos(\tau_a) = \sum_{a=1}^{k} (\breve{\b{x}}_a^\top \breve{\b{x}}_{a+1})/(||\breve{\b{x}}_a||_2 ||\breve{\b{x}}_{a+1}||_2)$ where $\breve{\b{x}}_a := \b{x}_a - \b{x}_i$ is the $a$-th edge of the polyhedron passing through the vertex $\b{x}_i$, $\b{x}_a$ is the $a$-th neighbor of $\b{x}_i$, and $\breve{\b{x}}_{a+1}$ denotes the next edge sharing the same polyhedron face with $\breve{\b{x}}_a$ where $\breve{\b{x}}_{k+1}=\breve{\b{x}}_1$. 

Note that finding the pairs of edges which belong to the same face is difficult and time-consuming so we relax this calculation to the summation of the cosine of angles between all pairs of edges meeting at the vertex $\b{x}_i$:
\begin{align}\label{equation_anomaly_score}
s_A(\b{x}_i) := \sum_{a=1}^{k-1} \sum_{b=a+1}^k \frac{\breve{\b{x}}_a^\top \breve{\b{x}}_b}{||\breve{\b{x}}_a||_2 ||\breve{\b{x}}_b||_2},
\end{align}
where $\breve{\b{x}}_a := \b{x}_a - \b{x}_i$, $\breve{\b{x}}_b := \b{x}_b - \b{x}_i$, and $\b{x}_a$ and $\b{x}_b$ denote the $a$-th and $b$-th neighbor of $\b{x}_i$. 
In Eq. (\ref{equation_anomaly_score}), we have omitted the redundant angles because of symmetry of inner product.
Note that the Eq. (\ref{equation_anomaly_score}) implies that we normalize the $k$ neighbors of $\b{x}_i$ to fall on the unit hyper-sphere centered at $\b{x}_i$ and then compute their cosine similarities (see Fig. \ref{figure_polyhedron}-c).

The mentioned relaxation is valid for the following reason. 
Take two edges meeting at the vertex $\b{x}_i$. If the two edges belong to the same polyhedron face, the relaxation is exact. Consider the case where the two edges do not belong to the same face. These two edges are connected with a set of polyhedron faces. If we tweak one of the two edges to increase/decrease the angle between them, the angle of that edge with its neighbor edge on the same face also increases/decreases. Therefore, the changes in the additional angles of relaxation are consistent with the changes of the angles between the edges sharing the same faces.

After scoring the data points, we can sort the points and find a suitable threshold visually using a scree plot of the scores. However, in order to find anomalies automatically, we apply K-means clustering, with two clusters, to the scores. The cluster with the larger mean is the cluster of anomalies because the higher the score, the more anomalous the point.

For finding anomalies for out-of-sample data, we find $k$-NN for the out-of-sample point where the neighbors are from the training points. Then, we calculate the anomaly score using Eq. (\ref{equation_anomaly_score}). The K-means cluster whose mean is closer to the calculated score determines whether the point is normal or anomaly. 
It is noteworthy that one can see anomaly detection for out-of-sample data as novelty detection \cite{pimentel2014review}.

\subsection{Kernel Curvature Anomaly Detection}

The pattern of normal and anomalous data might not be linear. Therefore, we propose \textit{Kernel CAD (K-CAD)} to work on data in the feature space.
In K-CAD, the two stages of finding $k$-NN and calculating the anomaly score are performed in the feature space. Let $\b{\phi}: \mathcal{X} \rightarrow \mathcal{H}$ be the pulling function mapping the data $\b{x} \in \mathcal{X}$ to the feature space $\mathcal{H}$. In other words, $\b{x} \mapsto \b{\phi}(\b{x})$. Let $t$ denote the dimensionality of the feature space, i.e., $\b{\phi}(\b{x}) \in \mathbb{R}^t$ while $\b{x} \in \mathbb{R}^d$. Note that we usually have $t \gg d$.
The kernel over two vectors $\b{x}_1$ and $\b{x}_2$ is the inner product of their pulled data \cite{hofmann2008kernel}: $\mathbb{R} \ni k(\b{x}_1, \b{x}_2) := \b{\phi}(\b{x}_1)^\top \b{\phi}(\b{x}_2)$.
The Euclidean distance in the feature space is \cite{scholkopf2001kernel}: $||\b{\phi}(\b{x}_i) - \b{\phi}(\b{x}_j)||_2 = \sqrt{k(\b{x}_i, \b{x}_i) -2 k(\b{x}_i, \b{x}_j) + k(\b{x}_j, \b{x}_j)}$.
Using this distance, we find the $k$-NN of the dataset in the feature space. 

After finding $k$-NN in the feature space, we calculate the score in the feature space. We pull the vectors $\breve{\b{x}}_a$ and $\breve{\b{x}}_b$ to the feature space so $\breve{\b{x}}_a^\top \breve{\b{x}}_b$ is changed to $k(\breve{\b{x}}_a, \breve{\b{x}}_b) = \b{\phi}(\breve{\b{x}}_a)^\top \b{\phi}(\breve{\b{x}}_b)$. 
Let $\b{K}_i \in \mathbb{R}^{k \times k}$ denote the kernel of neighbors of $\b{x}_i$ whose $(a,b)$-th element is $k(\breve{\b{x}}_a, \breve{\b{x}}_b)$. 
The vectors in Eq. (\ref{equation_anomaly_score}) are normalized. In the feature space, this is equivalent to normalizing the kernel $k(\breve{\b{x}}_a, \breve{\b{x}}_b) := k(\breve{\b{x}}_a, \breve{\b{x}}_b) / \sqrt{k(\breve{\b{x}}_a, \breve{\b{x}}_a)\, k(\breve{\b{x}}_b, \breve{\b{x}}_b)}$ \cite{ah2010normalized}.
If $\b{K}'_i$ denotes the normalized kernel $\b{K}_i$, the anomaly score in the feature space is:
\begin{align}\label{equation_anomaly_score_kernel}
s_A(\b{x}_i) := \sum_{a=1}^{k-1} \sum_{b=a+1}^k \b{K}'_i(a,b),
\end{align}
where $\b{K}'_i(a,b)$ is the $(a,b)$-th element of the kernel.
The K-means clustering and out-of-sample anomaly detection are similarly performed as in CAD.

Our observations in experiments showed that the anomaly score in K-CAD is ranked inversely for some kernels such as Radial Basis Function (RBF), Laplacian, and polynomial (different degrees) in various datasets. In other words, for example, in K-CAD with linear (i.e., CAD), cosine, and sigmoid kernels, the more anomalous points have greater score but in K-CAD with RBF, Laplacian, and polynomial kernels, the smaller score is assigned to the more anomalous points. We conjecture that the reason lies in the characteristics of the kernels. We defer more investigations for the reason as a future work.
In conclusion, for the mentioned kernels, we should either multiply the scores by $-1$ or take the K-means cluster with smaller mean as the anomaly cluster. 

\subsection{Anomaly Landscape and Anomaly Paths}

We define \textit{anomaly landscape} to be the landscape in the input space whose value at every point $\b{x}_i$ in the space is the anomaly score computed by Eq. (\ref{equation_anomaly_score}) or (\ref{equation_anomaly_score_kernel}).
The point $\b{x}_i$ in the space can be either the training or out-of-sample point but the $k$-NN is obtained from the training data. 
We can have two types of anomaly landscape where all the training data points or merely the non-anomaly training points are used for $k$-NN. In the latter type, the training phase of CAD or K-CAD are performed before calculating the anomaly landscape for the whole input space. 

We also define the \textit{anomaly path} as the path that an anomalous point has traversed from its not-known-yet normal version to become anomalous. Conversely, it is the path that an anomalous point should traverse to become normal.  
In other words, \emph{an anomaly path can be used to make a normal sample anomalous or vice-versa}.
At every point on the path, we calculate the $k$-NN again because the neighbors may change slightly during the path. For anomaly path, we use the second type of anomaly landscape where the path is like going up/down the mountains in this landscape. For finding the anomaly path for every anomaly point, we use gradient descent where the gradient of the Eq. (\ref{equation_anomaly_score}) is used:
\begin{equation}\label{equation_anomaly_score_gradient}
\begin{aligned}
\frac{\partial s_A(\b{x}_i)}{\partial \b{x}_i} &= \sum_{a=1}^{k-1} \sum_{b=a+1}^k \bigg[ \frac{1}{||\breve{\b{x}}_a||_2 ||\breve{\b{x}}_b||_2} \Big[\! -(\breve{\b{x}}_a + \breve{\b{x}}_b) + \breve{\b{x}}_a^\top \breve{\b{x}}_b \big(\frac{\breve{\b{x}}_a}{||\breve{\b{x}}_a||_2^2} + \frac{\breve{\b{x}}_b}{||\breve{\b{x}}_b||_2^2}\big) \Big] \bigg],
\end{aligned}
\end{equation}
whose derivation is eliminated for brevity (see supplementary material at the end of this article). 
The anomaly path can be computed in CAD and not K-CAD because the gradient in K-CAD cannot be computed analytically. 
The anomaly path can have many applications one of which is image denoising as explained in our experiments.

\section{Prototype Selection}

\subsection{Inverse Curvature Anomaly Detection}

If the anomaly detection uses scores, we can see instance ranking and numerosity reduction in the opposite perspective of anomaly detection. Therefore, the ranking scores can be considered as the anomaly scores multiplied by $-1$: $s_R(\b{x}_i) := -1 \times s_A(\b{x}_i) = -\sum_{a=1}^{k-1} \sum_{b=a+1}^k (\breve{\b{x}}_a^\top \breve{\b{x}}_b) / (||\breve{\b{x}}_a||_2 ||\breve{\b{x}}_b||_2)$.
We sort the ranking scores in descending order. The data point with larger ranking score is more important. As the order of ranking scores is inverse of the order of anomaly scores, we name this method as \textit{inverse CAD (iCAD)}. 

Prototype selection can be performed in two approaches: (I) the data points are sorted and a portion of the points having the best ranks is retained, or (II) a portion of data points is retained as prototypes and the rest of points are discarded. Some examples of the fist approach is IRMD, PSA, SOS, and SE. DROP3 and ENN are examples for the second approach. 
The iCAD can be used for both approaches. The first approach is ranking the points with the ranking score. 
For the second approach, we apply K-means clustering, with two clusters, to the ranking scores and take the points of the cluster with larger mean. 

\subsection{Kernel Inverse Curvature Anomaly Detection}

We can perform iCAD in the feature space to have \textit{Kernel iCAD (K-iCAD)}. The ranking score is again the anomaly score multiplied by $-1$ to reverse the ranks of scores: $s_R(\b{x}_i) := -1 \times s_A(\b{x}_i) = -\sum_{a=1}^{k-1} \sum_{b=a+1}^k \b{K}'_i(a,b)$.
Again, we have two approaches where the points are ranked or K-means is applied on the scores. 
Note that for what was mentioned before, we do not multiply by $-1$ for some kernels including RBF, Laplacian, and polynomial.  
Note that iCAD and K-iCAD are task agnostic and can be used for data reduction in classification, regression, and clustering. For classification, we apply the method for every class while in regression and clustering, the method is applied on the entire data.

\section{Experiments}

\subsection{Experiments for Anomaly Detection}

\begin{figure}[!t]
\makebox[\textwidth][c]{\includegraphics[width=1.2\textwidth]{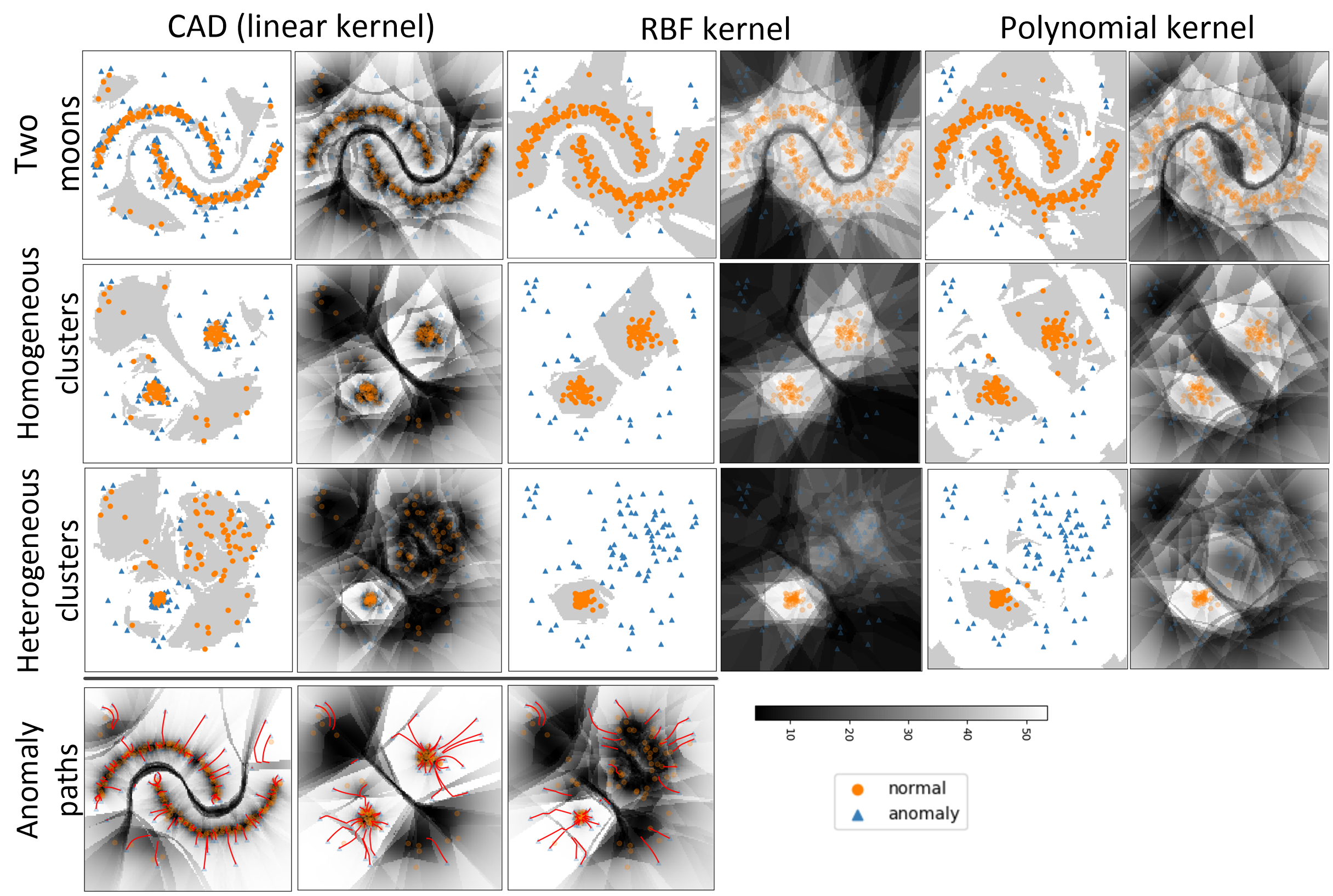}}%
\caption{Anomaly detection, anomaly scores, anomaly landscape, and anomaly paths for synthetic datasets. In the gray and white plots, the gray and white colors show the regions determined as normal and anomaly, respectively. The gray-scale plots are the anomaly scores.}
\label{figure_synthetic}
\end{figure}

\textbf{Synthetic Datasets:}
We examined CAD and iCAD on three two-dimensional synthetic datasets, i.e., two moons and two homogeneous and heterogeneous clusters. Figure \ref{figure_synthetic} shows the results for CAD and K-CAD with RBF and polynomial (degree three) kernels. As expected, the abnormal and core points are correctly detected as anomalous and normal points, respectively. The boundary points are detected as anomaly in CAD while they are correctly recognized as normal points in K-CAD. In heterogeneous clusters data, the larger cluster is correctly detected as normal in CAD but not in K-CAD; however, if the threshold is manually changed (rather than by K-means) in K-CAD, the larger cluster will also be correctly recognized. 
As seen in this figure, the scores are reverse in EBF and polynomial kernels which is consistent to our previous note in the paper. 
We also show the anomaly landscape and anomaly paths for CAD in Fig. \ref{figure_synthetic}. The K-CAD does not have anomaly paths as mentioned before. The landscapes in this figure are of the second type and the paths are shown by red traces which simulates climbing down the mountains in the landscape.

\noindent
\textbf{Real Datasets:}
We did experiments on several real datasets of anomaly detection. The datasets, which are taken from \cite{web_anomaly_datasets}, are speech, opt. digits, arrhythmia, wine, and musk with 1.65\%, 3\%, 15\%, 7.7\%, and 3.2\% portions of anomalies, respectively. The sample size of these datasets are 3686, 5216, 452, 129, and 3062 and their dimensionality are 400, 64, 274, 13, and 166, respectively. 
We compared CAD and K-CAD with RBF and polynomial (degree 3) kernels to Isolation forest, LOF, one-class SVM (RBF kernel), and EE. We used $k=10$ in LOF, CAD, and K-CAD. 
The average area under the ROC curve (AUC) and the average time for both training and test phases over 10-fold Cross Validation (CV) are reported in Table \ref{table_experiments_anomaly_detection}. For wine data, because of small sample size, we used 2-fold CV.
The system running the methods was Intel Core i7, 3.60 GHz, with 32 GB RAM.
In most cases, K-CAD has better performance than CAD; although CAD is useful and effective as we will see for anomaly path and also instance ranking. For speech and optdigits datasets, RBF kernel has better performance than polynomial and for other datasets, polynomial kernel is better. Mostly, K-CAD is faster in both training and test phases because K-CAD uses kernel matrix and normalizing the matrix rather than element-wise cosines in CAD. 
In speech and optdigits datasets, we outperform all the baseline methods in both trainign and test AUC rates. In arrhythmia data, K-CAD with polynomial kernel has beter results than isolation forest. For wine dataset, K-CAD with polynomial kernel is better than isolation forest, SVM, and EE. In musk data, K-CAD with both RBF and polynomial kernels is better than isolation forest and SVM.

\begin{figure}[!t]
\begin{floatrow}
\ffigbox{%
  
\includegraphics[width=2.2in]{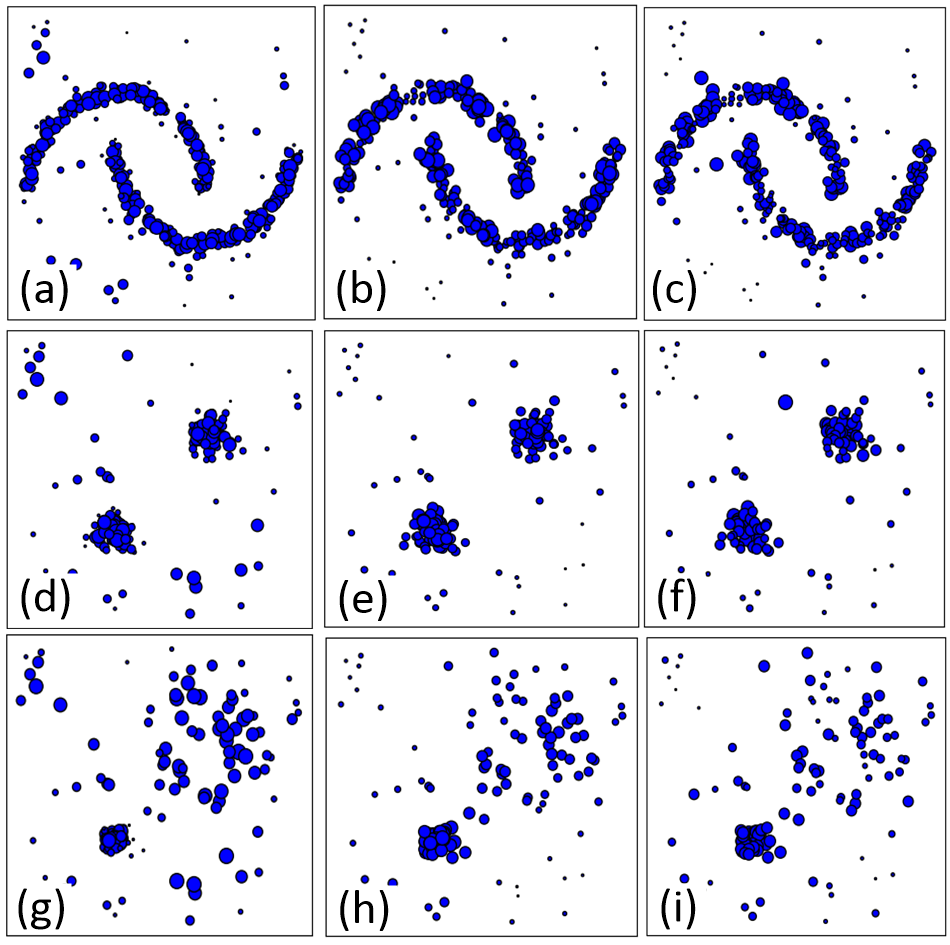}
  
}{%
  \caption{Instance ranking for synthetic datasets where larger markers are for more important data points. The first to third columns correspond to iCAD, K-iCAD with RBF kernel, and K-iCAD with polynomial kernel, respectively.}%
  \label{figure_synthetic2}
}
\capbtabbox{%
  
\scalebox{0.6}{    
\begin{tabular}{l | l | l | c | c | c}
\hline
\hline
\multicolumn{3}{c|}{} & \textbf{CAD} & \textbf{K-CAD (rbf)} & \textbf{K-CAD (poly)} \\ 
\hline
\multirow{4}{*}{$k$ = 3} 
& \multirow{2}{*}{ \textbf{Train:}} 
& \textbf{Time:} & 0.55 $\pm$ 0.00 & 3.25 $\pm$ 0.15 & 3.06 $\pm$ 0.16 \\
& & \textbf{AUC:} & 37.01 $\pm$ 1.00 & 49.83 $\pm$ 0.63 & 63.61 $\pm$ 1.36 \\
\cline{2-6}
& \multirow{2}{*}{ \textbf{Test:}} 
& \textbf{Time:} & 1.12 $\pm$ 0.02 & 0.37 $\pm$ 0.01 & 0.35 $\pm$ 0.01 \\
& & \textbf{AUC:} & 45.51 $\pm$ 6.21 & 49.32 $\pm$ 1.69 & 61.67 $\pm$ 10.18 \\
\hline
\multirow{4}{*}{$k$ = 10} 
& \multirow{2}{*}{ \textbf{Train:}} 
& \textbf{Time:} & 4.76 $\pm$ 0.02 & 2.75 $\pm$ 0.06 & 2.53 $\pm$ 0.03 \\
& & \textbf{AUC:} & 52.89 $\pm$ 0.96 & 48.87 $\pm$ 0.51 & 73.92 $\pm$ 1.12 \\
\cline{2-6}
& \multirow{2}{*}{ \textbf{Test:}} 
& \textbf{Time:} & 1.59 $\pm$ 0.01 & 0.30 $\pm$ 0.01 & 0.28 $\pm$ 00.00 \\
& & \textbf{AUC:} & 48.02 $\pm$ 9.06 & 48.56 $\pm$ 5.38 & 71.88 $\pm$ 9.23 \\
\hline
\multirow{4}{*}{$k$ = 20} 
& \multirow{2}{*}{ \textbf{Train:}} 
& \textbf{Time:} & 16.56 $\pm$ 0.07 & 3.41 $\pm$ 0.13 & 3.24 $\pm$ 0.24 \\
& & \textbf{AUC:} & 56.87 $\pm$ 1.02 & 49.33 $\pm$ 0.91 & 76.83 $\pm$ 1.07 \\
\cline{2-6}
& \multirow{2}{*}{ \textbf{Test:}} 
& \textbf{Time:} & 2.88 $\pm$ 0.01 & 0.39 $\pm$ 0.01 & 0.36 $\pm$ 0.01 \\
& & \textbf{AUC:} & 47.90 $\pm$ 9.18 & 49.27 $\pm$ 7.06 & 74.88 $\pm$ 8.47 \\
\hline
\hline
\end{tabular}%
}
  
}{%
  \caption{Comparison of CAD and K-CAD performance on arrhythmia dataset for different $k$ values.}%
  \label{table_experiments_anomaly_detection_k_value}
}
\end{floatrow}
\end{figure}

For experimenting the effect of $k$ in CAD and K-CAD, we report the results of $k \in \{3, 10, 20\}$ for arrhythmia dataset in Table \ref{table_experiments_anomaly_detection_k_value}.
For CAD, where the cosine is done element-wise, time increases by $k$ as expected. Overall, the accuracy, especially the training AUC, increases in CAD by $k$. K-CAD is more robust to change of $k$ in terms of accuracy and time. 

\begin{table*}[!t]
\begin{minipage}{\textwidth}
\caption{Comparison of anomaly detection methods. Rates are AUC percentage and times are in seconds.}
\label{table_experiments_anomaly_detection}
\renewcommand{\arraystretch}{1.3}  
\centering
\scalebox{0.6}{    
\begin{tabular}{l | l | l | c | c | c | c | c | c | c}
\hline
\hline
\multicolumn{3}{c|}{} & \textbf{CAD} & \textbf{K-CAD (rbf)} & \textbf{K-CAD (poly)} & \textbf{Iso Forest} & \textbf{LOF} & \textbf{SVM} & \textbf{EE} \\ 
\hline
\multirow{4}{*}{ \textbf{Speech}} 
& \multirow{2}{*}{ \textbf{Train:}} 
& \textbf{Time:} & 14.84 $\pm$ 0.21 & 13.54 $\pm$ 0.21 & 13.87 $\pm$ 0.33 & 2.82 $\pm$ 0.06 & 6.53 $\pm$ 0.02  & 6.12 $\pm$ 0.02 & 23.35 $\pm$ 0.25 \\
& & \textbf{AUC:} & 34.78 $\pm$ 0.15 & 76.15 $\pm$ 2.08 & 63.69 $\pm$ 1.48 & 48.37 $\pm$ 1.38 & 53.99 $\pm$ 1.75  & 46.63 $\pm$ 1.59 & 49.16 $\pm$ 1.84 \\
\cline{2-10}
& \multirow{2}{*}{ \textbf{Test:}} 
& \textbf{Time:} & 7.16 $\pm$ 0.03 & 1.38 $\pm$ 0.03 & 1.42 $\pm$ 0.03 & 0.06 $\pm$ 00.01 & 7.31 $\pm$ 0.10 & 0.24 $\pm$ 0.01 & 0.01 $\pm$ 0.00 \\
& & \textbf{AUC:} & 42.07 $\pm$ 10.48 & 71.23 $\pm$ 13.18 & 56.15 $\pm$ 10.48 & 45.23 $\pm$ 12.17 & 53.53 $\pm$ 12.29 & 44.55 $\pm$ 12.09 & 47.25 $\pm$ 12.54 \\
\hline
\hline
\multirow{4}{*}{ \textbf{Opt digits}} 
& \multirow{2}{*}{ \textbf{Train:}} 
& \textbf{Time:} & 13.27 $\pm$ 0.11 & 26.96 $\pm$ 0.31 & 25.86 $\pm$ 0.48 & 0.81 $\pm$ 0.01 & 1.84 $\pm$ 0.02  & 3.10 $\pm$ 0.01 & 0.96 $\pm$ 0.04 \\
& & \textbf{AUC:} & 32.67 $\pm$ 1.53 & 87.52 $\pm$ 1.76 & 77.79 $\pm$ 1.45 & 68.38 $\pm$ 4.64 & 60.84 $\pm$ 1.67  & 50.52 $\pm$ 3.81 & 39.04 $\pm$ 2.44 \\
\cline{2-10}
& \multirow{2}{*}{ \textbf{Test:}} 
& \textbf{Time:} & 11.24 $\pm$ 0.12 & 2.67 $\pm$ 0.01 & 2.59 $\pm$ 0.03 & 0.03 $\pm$ 0.01 & 2.13 $\pm$ 0.08 & 0.15 $\pm$ 00.00 & 00.01 $\pm$ 00.00 \\
& & \textbf{AUC:} & 26.28 $\pm$ 7.10 & 88.22 $\pm$ 5.62 & 79.72 $\pm$ 4.81 & 68.36 $\pm$ 8.11 & 61.12 $\pm$ 11.65 & 37.49 $\pm$ 7.41 & 38.84 $\pm$ 4.29 \\
\hline
\hline
\multirow{4}{*}{ \textbf{Arrhythmia}} 
& \multirow{2}{*}{ \textbf{Train:}} 
& \textbf{Time:} & 4.76 $\pm$ 0.02 & 2.75 $\pm$ 0.06 & 2.53 $\pm$ 0.03 & 0.20 $\pm$ 0.01 & 0.07 $\pm$ 00.00  & 0.13 $\pm$ 00.00 & 0.85 $\pm$ 0.02 \\
& & \textbf{AUC:} & 52.89 $\pm$ 0.96 & 48.87 $\pm$ 0.51 & 73.92 $\pm$ 1.12 & 62.43 $\pm$ 2.05 & 91.04 $\pm$ 0.66  & 88.56 $\pm$ 00.87 & 80.59 $\pm$ 0.65 \\
\cline{2-10}
& \multirow{2}{*}{ \textbf{Test:}} 
& \textbf{Time:} & 1.59 $\pm$ 0.01 & 0.30 $\pm$ 0.01 & 0.28 $\pm$ 00.00 & 0.02 $\pm$ 0.01 & 0.08 $\pm$ 0.00 & 0.01 $\pm$ 00.00 & 00.01 $\pm$ 00.00 \\
& & \textbf{AUC:} & 48.02 $\pm$ 9.06 & 48.56 $\pm$ 5.38 & 71.88 $\pm$ 9.23 & 63.07 $\pm$ 11.55 & 90.57 $\pm$ 5.47 & 90.03 $\pm$ 5.63 & 80.32 $\pm$ 4.73 \\
\hline
\hline
\multirow{4}{*}{ \textbf{Wine}} 
& \multirow{2}{*}{ \textbf{Train:}} 
& \textbf{Time:} & 0.28 $\pm$ 0.00 & 0.03 $\pm$ 0.03 & 0.03 $\pm$ 0.01 & 0.09 $\pm$ 0.00 & 0.01 $\pm$ 0.00  & 0.01 $\pm$ 0.00 & 0.05 $\pm$ 0.02 \\
& & \textbf{AUC:} & 25.59 $\pm$ 4.28 & 27.04 $\pm$ 10.66 & 92.11 $\pm$ 7.06 & 79.56 $\pm$ 10.59 & 98.70 $\pm$ 1.29  & 68.59 $\pm$ 4.25 & 59.56 $\pm$ 37.15 \\
\cline{2-10}
& \multirow{2}{*}{ \textbf{Test:}} 
& \textbf{Time:} & 0.18 $\pm$ 00.00 & 0.02 $\pm$ 00.00 & 0.01 $\pm$ 0.00 & 0.02 $\pm$ 0.00 & 0.01 $\pm$ 0.00 & 0.01 $\pm$ 0.00 & 0.01 $\pm$ 0.00 \\
& & \textbf{AUC:} & 23.65 $\pm$ 14.45 & 40.17 $\pm$ 13.12 & 86.97 $\pm$ 2.96 & 76.09 $\pm$ 10.11 & 92.58 $\pm$ 5.69 & 91.13 $\pm$ 3.83 & 57.70 $\pm$ 38.84 \\
\hline
\hline
\multirow{4}{*}{ \textbf{Musk}} 
& \multirow{2}{*}{ \textbf{Train:}} 
& \textbf{Time:} & 11.15 $\pm$ 0.20 & 9.67 $\pm$ 0.47 & 9.42 $\pm$ 0.03 & 0.98 $\pm$ 0.01 & 1.16 $\pm$ 0.03  & 3.16 $\pm$ 0.00 & 10.16 $\pm$ 0.32 \\
& & \textbf{AUC:} & 40.69 $\pm$ 2.97 & 69.68 $\pm$ 2.97 & 93.45 $\pm$ 1.46 & 99.91 $\pm$ 00.00 & 41.93 $\pm$ 3.34  & 57.99 $\pm$ 7.34 & 99.99 $\pm$ 00.00 \\
\cline{2-10}
& \multirow{2}{*}{ \textbf{Test:}} 
& \textbf{Time:} & 4.89 $\pm$ 0.02 & 0.98 $\pm$ 0.02 & 0.98 $\pm$ 0.01 & 0.03 $\pm$ 0.00 & 1.24 $\pm$ 0.01 & 0.17 $\pm$ 0.00 & 0.01 $\pm$ 0.00 \\
& & \textbf{AUC:} & 30.30 $\pm$ 10.37 & 50.00 $\pm$ 0.00 & 93.80 $\pm$ 3.77 & 99.95 $\pm$ 0.00 & 39.00 $\pm$ 10.55 & 5.71 $\pm$ 3.63 & 100 $\pm$ 0.00 \\
\hline
\hline
\end{tabular}%
}
\end{minipage}
\end{table*}

\begin{figure}[!t]
\makebox[\textwidth][c]{\includegraphics[width=1\textwidth]{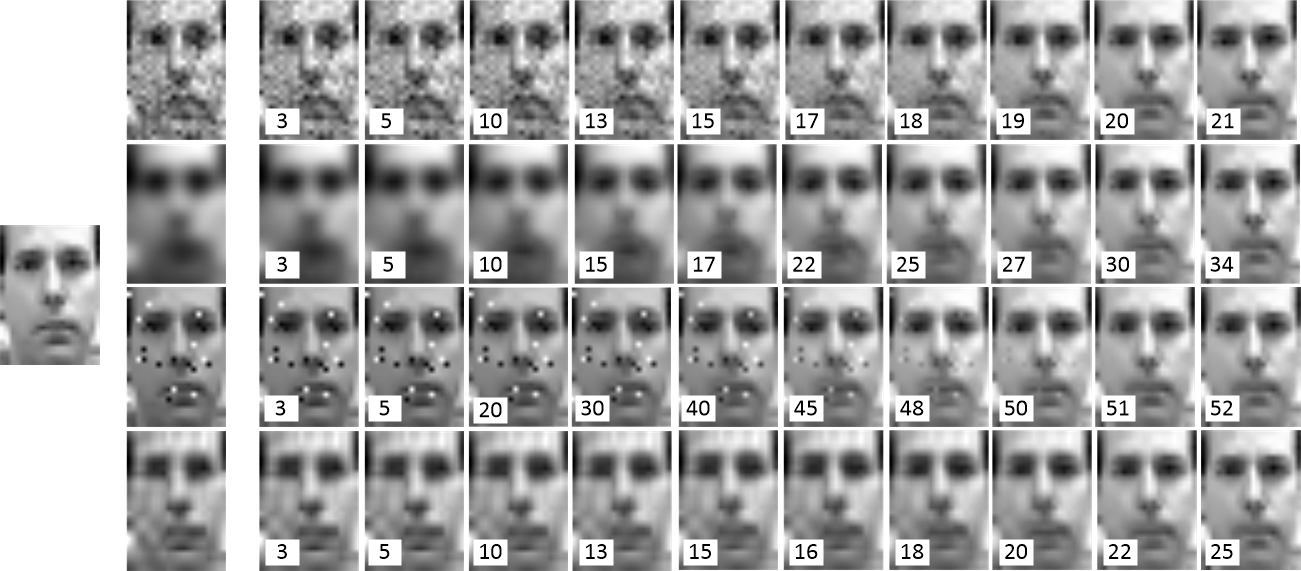}}%
\caption{Image denoising using anomaly paths: the most left image is the original image and the first to fourth rows are for Gaussian noise, Gaussian blurring, salt \& pepper impulse noise, and JPEG blocking. The numbers are the iteration indices.}
\label{figure_denoised_images}
\end{figure}

\noindent
\textbf{An Application; Image Denoising:}
One of the applications for anomaly path is image denoising where several similar reference images exist; for example, in video where neighbor frames exist. For experiment, we used the first 100 frames of Frey face dataset. We selected one of the frames and applied different types of noises, i.e., Gaussian noise, Gaussian blurring, salt \& pepper impulse noise, and JPEG blocking to it all with the same mean squared errors (MSE = $625$). For a more difficult experiment, we removed the non-distorted frame from dataset. Figure \ref{figure_denoised_images} shows the iterations of denoising for different noise types where $k=3$ is used.


\subsection{Experiments for Prototype Selection}

\textbf{Synthetic Datasets:}
The performances of iCAD and K-iCAD with RBF and polynomial (degree 3) kernels are illustrated in Fig. \ref{figure_synthetic2} for the three synthetic datasets where the larger markers show the more important points. We can see that the points are ranked as expected.

\noindent
\textbf{Real Datasets:}
We performed experiments on several real datasets, i.e., pima, image segment, Facebook metrics, and iris datasets, from the UCI machine learning repository. The first two datasets are used for classification, the third for regression, and the last one for clustering. The sample size of datasets are 768, 2310, 500, and 150, and their dimensionality are 8, 19, 19, and 4, respectively. The number of classes/clusters in pima, image segment, and iris are 2, 7, and 3, respectively. 
We used 1-Nearest Neighbor (1-NN), Linear Discriminant Analysis (LDA), SVM, Linear Regression (LR), Random Forest (RF), Multi-Layer Perceptron (MLP) with two hidden layers, K-means and Birch clustering methods in experiments.  
Table \ref{table_experiments_prototype_selection} reports the average accuracy and time over 10-fold CV and comparison to IRMD (with QR decomposition), PSA, SOS, Sorted by Distance from Mean (SDM), ENN, DROP3, and No Reduction (NR).

\begin{table*}[!t]
\begin{minipage}{\textwidth}
\caption{Comparison of instance selection methods in classification. Classification, regression, and clustering rates are accuracy, mean absolute error, and adjusted rand index (best is 1), respectively, and times are in seconds. The left and right columns are for rank-based and retaining-based methods. Numbers in parentheses show the percentage of retained data.}
\label{table_experiments_prototype_selection}
\renewcommand{\arraystretch}{1.3}  
\centering
\scalebox{0.55}{    
\begin{tabular}{l | l || c | c | c | c | c | c | c | c || c | c | c | c | c || c}
\hline
\hline
\multicolumn{16}{c}{\textbf{Pima Dataset}} \\
\hline
\hline
\multicolumn{2}{c||}{} & \textbf{iCAD} & \textbf{K-iCAD (rbf)} & \textbf{K-iCAD (poly)} & \textbf{IRMD} & \textbf{PSA} & \textbf{SOS} & \textbf{SE} & \textbf{SDM} & \textbf{iCAD} & \textbf{K-iCAD (rbf)} & \textbf{K-iCAD (poly)} & \textbf{ENN} & \textbf{DROP3} & \textbf{NR}\\ 
\hline
& \textbf{Time:} & 1.98E$+$0 & 4.41E$-$1 & 4.14E$-$1 & 9.52E$-$3 & 5.87E$+$0 & 2.13E$-$2 & 1.59E$-2$ & 4.51E$-$3 & 2.09E$+$0 & 4.73E$-$1 & 4.73E$-$1 & 1.87E$-$2 & 5.00E$+$0 & $\times$ \\
\hline
\multirow{3}{*}{ \textbf{1NN}} 
& 20\% data: & 70.69\% & 67.70\% & 61.97\% & 67.28\%  & 66.53\% & 67.06\% & 44.40\% & 62.87\% & (40.79\%) & (4.35\%) & (12.76\%) & (69.40\%) & (13.44\%) & \multirow{3}{*}{68.48\%}\\
& 50\% data: & 70.83\% & 66.01\% & 64.96\% & 66.64\%  & 64.44\% & 67.04\% & 50.64\% & 66.39\% & 65.23\% & 66.93\% & 65.88\% & 71.74\% & 64.06\% &  \\
& 70\% data: & 69.26\% & 67.56\% & 65.74\% & 67.95\%  & 65.75\% & 66.91\%  & 65.62\% & 67.81\% & & & & & &  \\
\hline
\hline
\multirow{3}{*}{ \textbf{LDA}}
& 20\% data: & 75.64\% & 67.82\% & 72.25\% & 70.81\%  & 76.17\% & 73.56\%  & 54.55\% & 65.09\% & & & & & &  \multirow{3}{*}{77.73\%}\\
& 50\% data: & 76.42\% & 76.03\% & 76.43\% & 73.81\%  & 76.81\% & 76.43\% & 71.35\% & 73.68\% & 65.62\% & 71.74\% & 66.14\% & 77.07\% & 75.12\% & \\
& 70\% data: & 76.94\% & 76.82\% & 76.81\% & 75.38\%  & 76.82\% & 76.56\%  & 76.55\% & 74.98\% & & & & & & \\
\hline
\hline
\multirow{3}{*}{ \textbf{SVM}}  
& 20\% data: & 63.28\% & 57.65\% & 62.50\% & 59.62\%  & 57.68\% & 63.52\% & 42.04\% & 63.78\% & & & & & &  \multirow{3}{*}{64.57\%}\\
& 50\% data: & 65.22\% & 62.25\% & 57.94\% & 61.57\%  & 61.57\% & 62.25\%  & 48.15\% & 60.01\% & 64.84\% & 63.92\% & 63.93\% & 67.18\% & 53.12\% &  \\
& 70\% data: & 62.76\% & 61.28\% & 56.40\% & 55.97\%  & 60.42\% & 62.63\% & 55.23\% & 64.98\% & & & & & & \\
\hline
\hline
\multicolumn{16}{c}{\textbf{Image Segment dataset}} \\
\hline
\hline
\multicolumn{2}{c||}{} & \textbf{iCAD} & \textbf{K-iCAD (rbf)} & \textbf{K-iCAD (poly)} & \textbf{IRMD} & \textbf{PSA} & \textbf{SOS} & \textbf{SE} & \textbf{SDM} & \textbf{iCAD} & \textbf{K-iCAD (rbf)} & \textbf{K-iCAD (poly)} & \textbf{ENN} & \textbf{DROP3} & \textbf{NR}\\ 
\hline
& \textbf{Time:} & 6.25E$+$0 & 1.20E$+$0  & 1.04E$+$0 & 3.64E$-$2 & 3.28E$+$2 & 5.86E$-$2 & 4.85E$-$2 & 1.38E$-$2 & 6.13E$+$0 & 1.58E$+$0 & 1.44E$+$0 & 1.30E$-$1 & 3.13E$+$2 & $\times$ \\
\hline
\multirow{3}{*}{ \textbf{1NN}} 
& 20\% data: & 90.34\% & 81.42\% & 83.72\% & 78.00\% & 90.25\% & 89.26\% & 83.41\% & 80.99\% & (9.54\%) & (3.97\%) & (3.14\%) & (95.25\%) & (6.78\%) &  \multirow{3}{*}{96.45\%}\\
& 50\% data: & 93.41\% & 89.43\% & 92.85\% & 86.83\%  & 94.76\% & 94.19\% & 84.71\% & 87.05\% & 47.22\% & 49.13\% & 49.43\% & 94.45\% & 60.34\% & \\
& 70\% data: & 94.84\% & 92.20\% & 95.75\% & 91.03\%  & 96.06\% & 95.41\% & 90.69\% & 90.00\% & & & & & & \\
\hline
\hline
\multirow{3}{*}{ \textbf{LDA}}
& 20\% data: & 89.26\% & 85.97\% & 90.86\% & 82.90\%  & 90.47\% & 90.51\% & 85.23\% & 86.36\% & & & & & &  \multirow{3}{*}{91.55\%}\\
& 50\% data: & 90.64\% & 90.04\% & 91.94\% & 86.32\%  & 91.08\% & 91.16\% & 86.27\% & 87.83\% & 60.60\% & 58.70\% & 59.91\% & 67.57\% & 79.35\% & \\
& 70\% data: & 90.90\% & 90.99\% & 91.64\% & 88.26\%  & 91.42\% & 91.34\% & 90.30\% & 89.48\% & & & & & & \\
\hline
\hline
\multirow{3}{*}{ \textbf{SVM}}  
& 20\% data: & 73.41\% & 77.74\% & 76.40\% & 71.42\%  & 78.44\% & 78.35\% & 71.73\% & 71.86\% & & & & & &  \multirow{3}{*}{85.23\%}\\
& 50\% data: & 80.90\% & 81.08\% & 80.30\% & 80.60\%  & 77.18\% & 82.85\% & 70.60\% & 81.34\% & 39.43\% & 40.38\% & 38.70\% & 78.52\% & 55.10\% & \\
& 70\% data: & 84.71\% & 84.71\% & 83.67\% & 78.87\%  & 82.20\% & 85.67\% & 86.79\% & 84.32\% & & & & & & \\
\hline
\hline
\multicolumn{16}{c}{\textbf{Facebook Dataset}} \\
\hline
\hline
\multicolumn{2}{c||}{} & \textbf{iCAD} & \textbf{K-iCAD (rbf)} & \textbf{K-iCAD (poly)} & \textbf{IRMD} & \textbf{PSA} & \textbf{SOS} & \textbf{SE} & \textbf{SDM} & \textbf{iCAD} & \textbf{K-iCAD (rbf)} & \textbf{K-iCAD (poly)} & \textbf{ENN} & \textbf{DROP3} & \textbf{NR}\\ 
\hline
& \textbf{Time:} & 1.30E$+$0 & 6.33E$-$1 & 1.01E$+$0 & 7.01E$-$3 & 1.08E$+$0 & 1.40E$-$2 & 8.40E$-$3 & 4.81E$-$3 & 1.26E$+$0 & 3.18E$-$1 & 3.18E$-$1 & $\times$ & $\times$ & $\times$ \\
\hline
\multirow{3}{*}{ \textbf{LR}} 
& 20\% data: & 7.57E$+$3 & 6.00E$+$3 & 7.45E$+$3 & 5.28E$+$3  & 9.98E$+$3 & 6.07E$+$3  & 1.89E$+$4 & 7.50E$+$3 & (49.96\%) & (6.95\%) & (34.81\%) & & & \multirow{3}{*}{5.81E$+$3} \\
& 50\% data: & 6.11E$+$3 & 5.51E$+$3 & 5.86E$+$3 & 4.85E$+$3  & 6.30E$+$3  & 6.25E$+$3  & 6.59E$+$3 & 5.69E$+$3 & 6.10E$+$3 & 1.56E$+$4 & 6.41E$+$3 & $\times$ & $\times$ & \\
& 70\% data: & 5.70E$+$3 & 6.00E$+$3 & 5.72E$+$3 & 4.95E$+$3  & 5.55E$+$3 & 5.90E$+$3  & 5.83E$+$3 & 5.30E$+$3 & & & & & & \\
\hline
\hline
\multirow{3}{*}{ \textbf{RF}}
& 20\% data: & 8.54E$+$3 & 7.87E$+$3 & 6.85E$+$3 & 5.53E$+$3  & 7.12E$+$3  & 6.56E$+$3  & 1.09E$+$4 & 6.58E$+$3 & & & & & & \multirow{3}{*}{6.17E$+$3} \\
& 50\% data: & 6.41E$+$3 & 7.08E$+$3 & 6.28E$+$3 & 5.02E$+$3  & 6.20E$+$3 & 6.76E$+$3  & 6.10E$+$3 & 7.19E$+$3 & 6.36E$+$3 & 9.82E$+$3 & 6.46E$+$3 & $\times$ & $\times$ & \\
& 70\% data: & 5.92E$+$3 & 6.95E$+$3 & 6.19E$+$3 & 5.03E$+$3  & 5.86E$+$3 & 6.26E$+$3  & 6.32E$+$3 & 6.03E$+$3 & & & & & & \\
\hline
\hline
\multirow{3}{*}{ \textbf{MLP}}
& 20\% data: & 1.348E$+$4 & 1.61E$+$4 & 1.02E$+$4 & 7.11E$+$3  & 2.12E$+$4 & 7.46E$+$3  & 4.70E$+$4 & 2.31E$+$4 & & & & & & \multirow{3}{*}{5.72E$+$3} \\
& 50\% data: & 6.10E$+$3 & 5.44E$+$3 & 5.57E$+$3 & 4.75E$+$3  & 6.93E$+$3 & 6.02E$+$3  & 5.62E$+$3 & 6.64E$+$3 & 6.11E$+$3 & 5.06E$+$4 & 8.31E$+$3 & $\times$ & $\times$ & \\
& 70\% data: & 6.31E$+$3 & 6.06E$+$3 & 5.66E$+$3 & 5.14E$+$3  & 5.75E$+$3 & 5.95E$+$3 & 5.86E$+$3 & 6.00E$+$3 & & & & & & \\
\hline
\hline
\multicolumn{16}{c}{\textbf{Iris Dataset}} \\
\hline
\hline
\multicolumn{2}{c||}{} & \textbf{iCAD} & \textbf{K-iCAD (rbf)} & \textbf{K-iCAD (poly)} & \textbf{IRMD} & \textbf{PSA} & \textbf{SOS} & \textbf{SE} & \textbf{SDM} & \textbf{iCAD} & \textbf{K-iCAD (rbf)} & \textbf{K-iCAD (poly)} & \textbf{ENN} & \textbf{DROP3} & \textbf{NR}\\ 
\hline
& \textbf{Time:} & 4.96E$-$1 & 5.30E$-$2 & 4.46E$-$2 & 1.60E$-$3 & 2.83E$-$1 & 3.70E$-$3 & 3.00E$-$1 & 1.90E$-$3 & 4.68E$-$1 & 6.07E$-$2 & 5.95E$-$2 & $\times$ & $\times$ & $\times$ \\
\hline
\multirow{3}{*}{ \textbf{K-means}} 
& 20\% data: & 6.87E$-$1 & 1.56E$-$1 & 2.34E$-$1 & 1.32E$-$1  & 6.18E$-$1 & 6.92E$-$1  & 5.02E$-$1 & 4.89E$-$1 & (60.59\%) & (84.07\%) & (84.07\%) & & & \multirow{3}{*}{7.03E$-$1}\\
& 50\% data: & 6.98E$-$1 & 7.13E$-$1 & 8.33E$-$1 & 5.93E$-$1  & 7.01E$-$1  & 7.38E$-$1  & 4.85E$-$1 & 5.77E$-$1 & 7.34E$-$1 & 8.06E$-$1 & 8.20E$-$1 & $\times$ & $\times$ \\
& 70\% data: & 7.18E$-$1 & 7.83E$-$1 & 8.20E$-$1 & 6.53E$-$1  & 7.18E$-$1 & 7.35E$-$1  & 6.27E$-$1 & 7.12E$-$1 & & & & & \\
\hline
\hline
\multirow{3}{*}{ \textbf{Birch}}
& 20\% data: & 6.09E$-$1 & 0.00E$+$0 & 1.26E$-$1 & 1.07E$-$1  & 6.97E$-$1 & 6.41E$-$1  & 5.10E$-$1 & 2.90E$-$1 & & & & & & \multirow{3}{*}{5.93E$-$1}\\
& 50\% data: & 6.48E$-$1 & 6.58E$-$1 & 7.15E$-$1 & 6.35E$-$1  & 6.11E$-$1 & 5.34E$-$1  & 5.11E$-$1 & 5.92E$-$1 & 7.71E$-$1 & 6.60E$-$1 & 6.85E$-$1 & $\times$ & $\times$ \\
& 70\% data: & 6.67E$-$1 & 7.17E$-$1 & 7.37E$-$1 & 6.61E$-$1  & 6.48E$-$1 & 5.79E$-$1  & 6.23E$-$1 & 6.04E$-$1 & & & & & \\
\hline
\hline
\end{tabular}%
}
\end{minipage}
\end{table*}

The iCAD and K-iCAD are reported in both rank-based and retaining-based versions of prototype selection. 
For pima and image segment datasets, iCAD and K-iCAD are both performing equally well but in other datasets, K-iCAD is mostly better. 
In terms of time, we outperform PSA and DROP3. 
In pima, we outperform all other baselines. In image segment, we are better than IRMD, SE, and SDM. In facebook data, we are mostly better than SOS, SE, and SDM, and in some cases better than PSA. In iris data, we outperform all the baselines. 
In some cases, we even outperform using the entire data. 
In retaining-based iCAD and K-iCAD, mostly, K-iCAD with RBF kernel retains the least, then K-iCAD with polynomial kernel, and then CAD. 

\section{Conclusion and Future Direction}

This paper proposed a new method for anomaly detection, named CAD, and its kernel version. The main idea was to consider every point as a vertex of a polyhedron with the help of its neighbors and measure its curvature. Moreover, with an opposite view to CA, iCAD and K-iCAD were proposed for prototype selection. Different experiments as well as an application in image denoising were also reported. As a possible future work, we will try the idea of curvature for manifold embedding to propose a curvature preserving embedding method.

\bibliographystyle{splncs}      
\bibliography{references.bib}            

\section*{Supplementary Material: Proof of Eq. (\ref{equation_anomaly_score_gradient})}

Let $f := \alpha / \beta := \breve{\b{x}}_a^\top \breve{\b{x}}_b / (||\breve{\b{x}}_a||_2 ||\breve{\b{x}}_b||_2)$. Then, $\beta^2 = ||\breve{\b{x}}_a||_2^2 ||\breve{\b{x}}_b||_2^2 = \breve{\b{x}}_a^\top \breve{\b{x}}_a \breve{\b{x}}_b^\top \breve{\b{x}}_b = (\b{x}_a^\top \b{x}_a - 2 \b{x}_a^\top \b{x}_i + \b{x}_i^\top \b{x}_i) (\b{x}_b^\top \b{x}_b - 2 \b{x}_b^\top \b{x}_i + \b{x}_i^\top \b{x}_i)$.
\begin{align*}
&\frac{\partial \beta^2}{\partial \b{x}_i} = (-2\b{x}_a + 2 \b{x}_i) ||\breve{\b{x}}_b||_2^2 + (-2\b{x}_b + 2 \b{x}_i) ||\breve{\b{x}}_a||_2^2 = -2 ||\breve{\b{x}}_b||_2^2\, \breve{\b{x}}_a -2 ||\breve{\b{x}}_a||_2^2\, \breve{\b{x}}_b. \\
&\frac{\partial \beta^2}{\partial \b{x}_i} = \frac{\partial \beta^2}{\partial \beta} \times \frac{\partial \beta}{\partial \b{x}_i} = 2 \beta \frac{\partial \beta}{\partial \b{x}_i} \implies \frac{\partial \beta}{\partial \b{x}_i} = \frac{1}{2 \beta} \frac{\partial \beta^2}{\partial \b{x}_i} \\
&\implies \frac{\partial \beta}{\partial \b{x}_i} = -\frac{||\breve{\b{x}}_b||_2}{||\breve{\b{x}}_a||_2} \breve{\b{x}}_a - \frac{||\breve{\b{x}}_a||_2}{||\breve{\b{x}}_b||_2} \breve{\b{x}}_b. \\
&\frac{\partial f}{\partial \b{x}_i} = \frac{\partial}{\partial \b{x}_i} (\frac{\alpha}{\beta}) = \frac{1}{\beta} \Big[\frac{\partial \alpha}{\partial \b{x}_i} - f \frac{\partial \beta}{\partial \b{x}_i}\Big]. \\
&\frac{\partial \alpha}{\partial \b{x}_i} = \frac{\partial (\breve{\b{x}}_a^\top \breve{\b{x}}_b)}{\partial \b{x}_i} = \frac{\partial}{\partial \b{x}_i} \big( (\b{x}_a - \b{x}_i)^\top (\b{x}_b - \b{x}_i) \big) = 2\b{x}_i - \b{x}_a - \b{x}_b = -( \breve{\b{x}}_a + \breve{\b{x}}_b ). \\
&f \frac{\partial \beta}{\partial \b{x}_i} = \frac{\breve{\b{x}}_a^\top \breve{\b{x}}_b}{||\breve{\b{x}}_a||_2 ||\breve{\b{x}}_b||_2} \Big[ -\frac{||\breve{\b{x}}_b||_2}{||\breve{\b{x}}_a||_2} \breve{\b{x}}_a - \frac{||\breve{\b{x}}_a||_2}{||\breve{\b{x}}_b||_2} \breve{\b{x}}_b \Big] = -\breve{\b{x}}_a^\top \breve{\b{x}}_b \big(\frac{\breve{\b{x}}_a}{||\breve{\b{x}}_a||_2^2} + \frac{\breve{\b{x}}_b}{||\breve{\b{x}}_b||_2^2}\big). \\
&\therefore ~~ \frac{\partial f}{\partial \b{x}_i} = \frac{1}{||\breve{\b{x}}_a||_2 ||\breve{\b{x}}_b||_2} \Big[ -(\breve{\b{x}}_a + \breve{\b{x}}_b) + \breve{\b{x}}_a^\top \breve{\b{x}}_b \big(\frac{\breve{\b{x}}_a}{||\breve{\b{x}}_a||_2^2} + \frac{\breve{\b{x}}_b}{||\breve{\b{x}}_b||_2^2}\big) \Big]. \\
&\frac{\partial s_A(\b{x}_i)}{\partial \b{x}_i} = \sum_{a=1}^{k-1} \sum_{b=a+1}^k \frac{\partial f_x}{\partial \b{x}_i},
\end{align*}
which gives the proposed derivative. Q.E.D.

\end{document}